\documentclass[letterpaper]{article} %
\usepackage{spconf,amsmath,graphicx}
\usepackage{epsfig}
\usepackage{graphicx}
\usepackage{enumitem}
\usepackage{amsmath}
\usepackage{amssymb}
\usepackage[font={small,it}]{caption}
\usepackage{subcaption}
\usepackage[ruled]{algorithm}
\usepackage[noend]{algpseudocode}
\usepackage{lineno}
\usepackage{booktabs}
\usepackage{multirow}
\usepackage{placeins}
\usepackage{array,booktabs}
\usepackage[table]{xcolor}
\newcolumntype{L}{@{}>{\kern\tabcolsep}l<{\kern\tabcolsep}}
\usepackage{colortbl}
\usepackage{xcolor}
\usepackage[T1]{fontenc}
\usepackage[utf8]{inputenc}
\usepackage{float}

\title{Context Driven Label Fusion for Segmentation of Subcutaneous and Visceral Fat in CT Volumes}
\name{Sarfaraz Hussein$^1$, Aileen Green$^2$, Arjun Watane$^1$, Georgios Papadakis$^3$, Medhat Osman$^2$, Ulas Bagci$^1$}
\address{$^1$Center for Research in Computer Vision (CRCV), University of Central Florida (UCF) \\ 
$^2$Department of Radiology, St. Louis University, St. Louis, MO \\
$^3$Radiology and Imaging sciences, National Institutes of Health (NIH), Bethesda MD}

\begin{document}

\maketitle
\begin{abstract}

Quantification of adipose tissue (fat) from computed tomography (CT) scans is conducted mostly through manual or semi-automated image segmentation algorithms with limited efficacy. In this work, we propose a completely unsupervised and automatic method to identify adipose tissue, and then separate Subcutaneous Adipose Tissue (SAT) from Visceral Adipose Tissue (VAT) at the abdominal region. We offer a three-phase pipeline consisting of (1) Initial boundary estimation using gradient points, (2) boundary refinement using Geometric Median Absolute Deviation and Appearance based Local Outlier Scores (3) Context driven label fusion using Conditional Random Fields (CRF) to obtain the final boundary between SAT and VAT. We evaluate the proposed method on 151 abdominal CT scans and obtain state-of-the-art 94\% and 91\% dice similarity scores for SAT and VAT segmentation, as well as significant reduction in fat quantification error measure.
\end{abstract}
\begin{keywords}
Fat Segmentation, CT, Conditional Random Fields, Subcutaneous fat, Visceral fat
\end{keywords}
\section{Introduction}
\label{sec:intro}
Quantification of adipose tissue (i.e., fat) and its subtypes is an important task in many clinical applications such as obesity, cardiac, and diabetes research. Among them, obesity is one of the most prevalent health conditions in recent years \cite{ng2014global}. In the United States, about one-third of the adult population is obese \cite{ogden2014prevalence}, causing an increased risk for cardiovascular diseases, diabetes, and certain types of cancer. Traditionally, Body Mass Index (BMI) has been used as a measure of obesity; however, it remains inconsistent across subjects, especially for underweight and obese individuals. Instead, volumetry of visceral adipose tissue from CT volumes is considered as a reliable, accurate and consistent measure of body fat distribution and its extent. However, current radiological quantification methods are insufficient, often requiring manual interaction at several anatomical locations, leading to inefficient and inaccurate quantification. 

\begin{figure}[t!]
\centering
\includegraphics[height=5.0 cm, width=80mm]{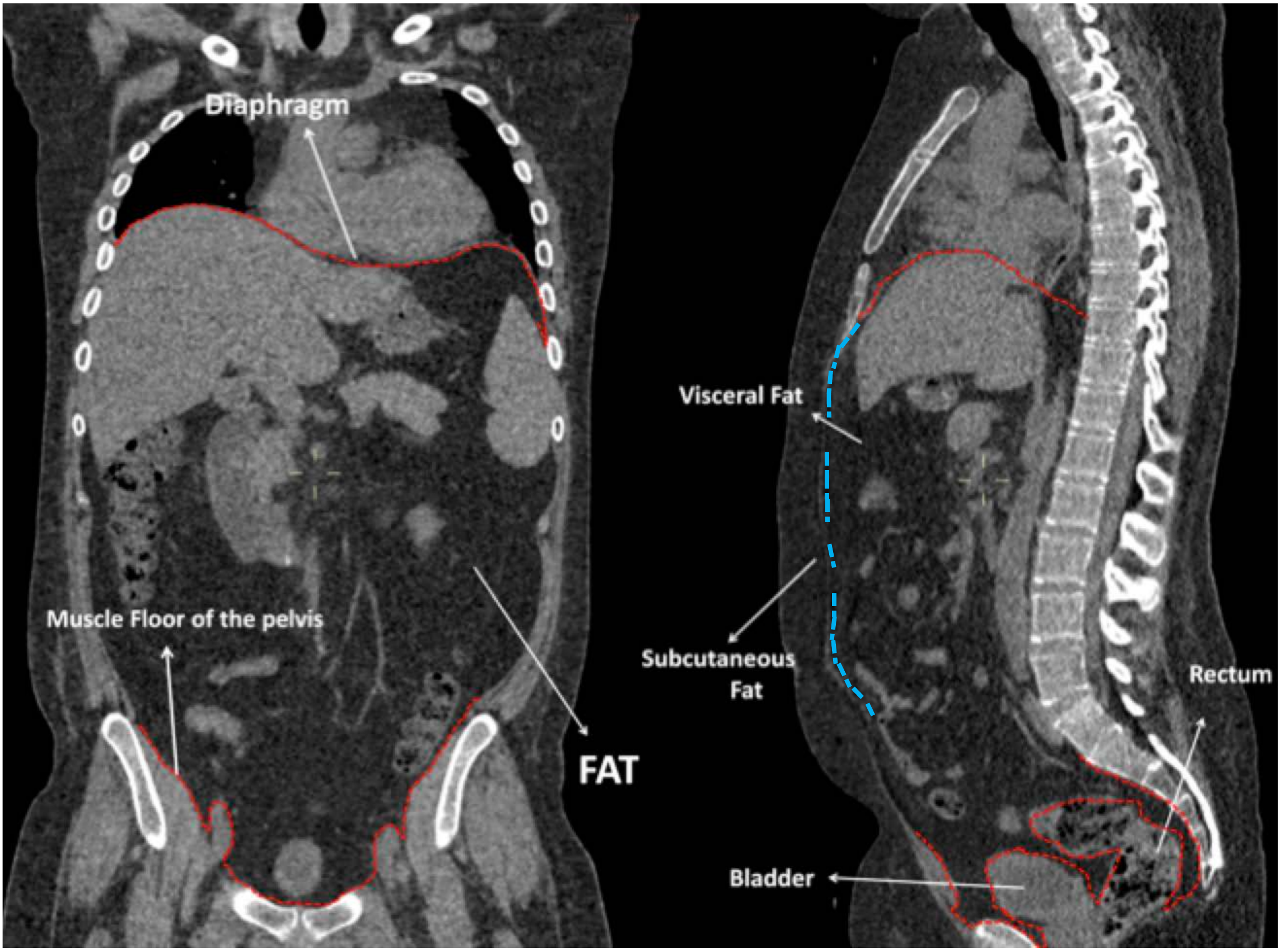}
\caption{\textit{Dotted red lines show the abdominal region boundaries on the left. A thin discontinuous muscle wall (dotted blue curve) separating VAT and SAT is shown on the right.}}
\label{fig:subvisfig}
\vspace{- 0.5 cm}
\end{figure}

Compared to SAT, body composition phenotype due to VAT is associated with medical disorders such as coronary heart disease, and several malignancies including prostate, breast and colorectal cancers. Hence, quantification of visceral/abdominal obesity is vital for precise diagnosis and timely treatment. However, separation of VAT from SAT is not trivial because both SAT and VAT regions share similar intensities in CT, and are vastly connected (See Figure \ref{fig:subvisfig}). Radiologists often rely on different morphological and filtering operators to segregate these two fat types in routine evaluations, but the size of structural element and neighborhood area for filtering are subjective and require excessive manual tuning.

In this work, we propose a novel method to identify, segment as well as quantify SAT and VAT automatically. Our contributions in this paper are the following. (1) Our proposed automated method is completely unsupervised and we estimate SAT-VAT separation boundary using both appearance and geometric cues. (2) The contextual information captured by our method can well handle inter-subject variations as compared to other methods. To the best of our knowledge, this is the largest VAT and SAT segmentation study (over 150 CT scans) till date validating an automated fat segmentation and quantification method.\\

\noindent \textbf{Related Work:} Body fat quantification has been a long-time active area of research for medical imaging scientists. For abdominal fat quantification, Zhao et al. \cite{zhao2006} used intensity profile along the radii connecting sparse points on the outer wall (skin boundary) starting from the abdominal body center. Boundary contour is then refined by a smoothness constraint to separate VAT from SAT. This method, however, does not adapt to obese patients easily where the neighboring subcutaneous and/or visceral fat cavities lead to leakage in segmentation. In another study, Romero et al. \cite{romero2006} developed a semi-supervised method generating skin-boundary, abdominal (visceral) wall, and subcutaneous fat masks. In a similar fashion, the method in \cite{pednekar2005automatic} is based on a hierarchical fuzzy affinity function based semi-supervised segmentation. Its success was vague when patient specific quantification is considered. In more advanced method of \cite{chung2009automated}, SAT, VAT and muscle are separated using a joint shape and appearance model. Mensink et al. \cite{mensink2011development} proposed a series of morphological operations but fine tuning of the algorithm was a necessary step for patient specific quantification, and this step should be repeated almost for every patient when the abdominal wall is too thin.  More recently, Kim et al. \cite{kim2013body} generated subcutaneous fat mask using a modified ``AND'' operation on four different directed masks with some success shown. However, logical and morphological operations make the whole quantification system vulnerable to inefficiencies. 

In comparison to all these methods, our proposed framework is robust to handle \textbf{patient specific variations} since we use both appearance and geometric information to generate the muscle boundary. Moreover, our proposed label fusion method using conditional random field (CRF) helps capture the contextual and pairwise similarity between image attributes leading to significantly better segmentation of SAT and VAT.   

\begin{figure}[t!]
\centering
\includegraphics[width=90mm]{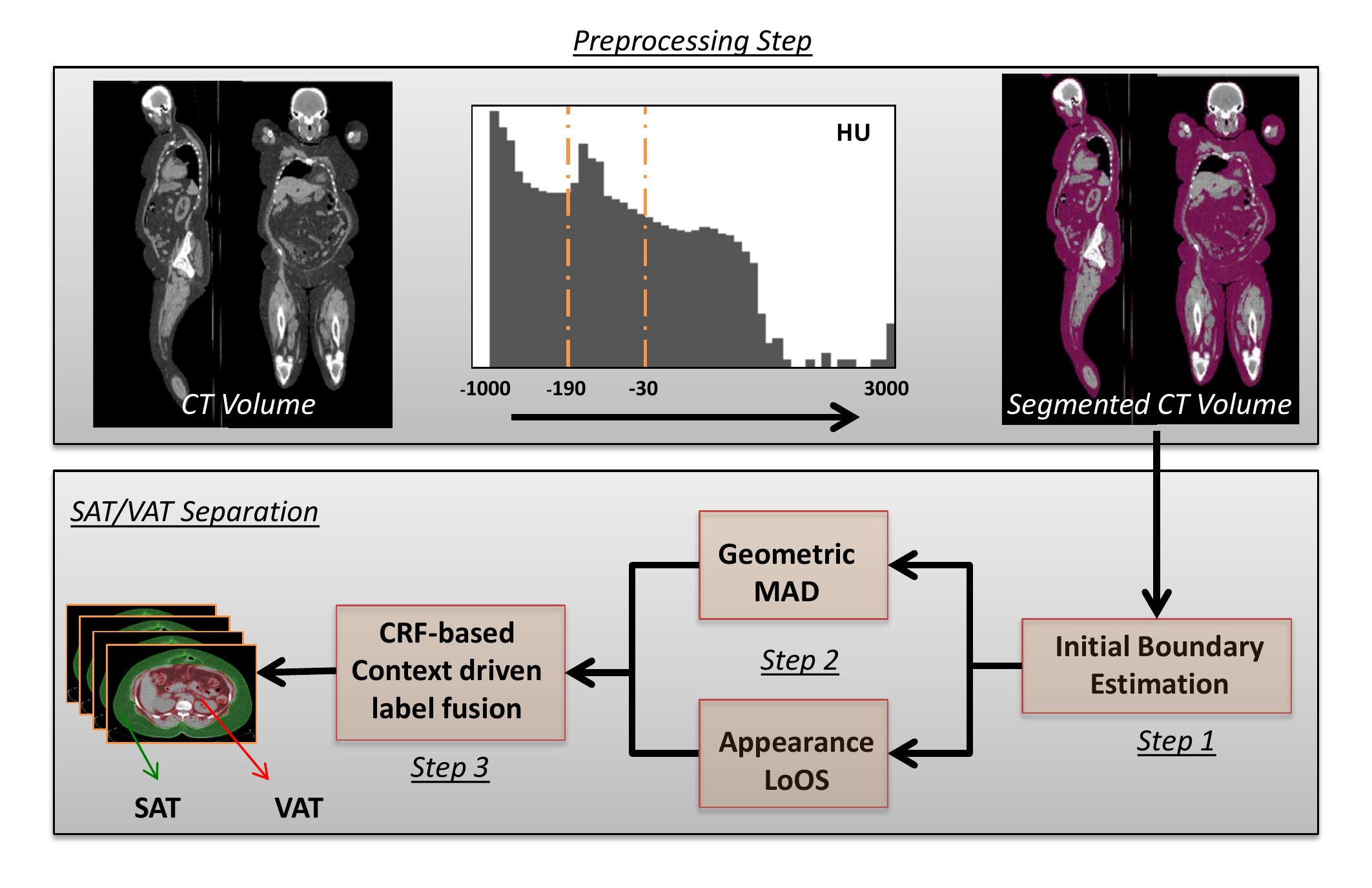}
\caption{\textit{Workflow of the proposed method. Upper: pre-processing step. Bottom: two stage outlier removal using MAD and LoOS and context driven label fusion via CRF.}}
\label{fig:workflow}
\vspace{-0.4 cm}
\end{figure}

\section{Methods}
\label{sec:Method}
The motivation for the proposed approach is based on the observation that in CT volumes, SAT and VAT are separated by a thin layer of non-fat tissue (Figure~\ref{fig:subvisfig}). Our proposed algorithm includes a pre-processing task and a three-step segmentation framework as illustrated in Figure~\ref{fig:workflow}. Since Hounsfield Unit (HU) interval for certain tissues in CT image is fixed, it is straightforward to identify whole-body fat regions from CT scans using a thresholding interval on HU space. In pre-processing step, we apply an identical set of morphological operations to remove noise and normalize the imaging data. Afterwards, in the first step of the core segmentation method, we identify the initial boundaries for VAT and SAT regions by using a sparse search conducted over a line oriented towards the center of the abdominal cavity starting from skin boundary. In the second step, we propose to use median absolute deviation (MAD) and local outlier scores (LoOS) in order to remove false positive boundary locations and thus refine VAT and SAT separation. In the final step, we utilize a label fusion framework using CRF energy minimization technique combining shape, anatomy, and appearance information.

\subsection{Pre-processing of CT Images}

The input to our pipeline is a CT abdominal volume. We begin by thresholding the CT volumes between -190 to -30 HU, corresponding to the fat tissue~\cite{yoshizumi1999abdominal}. For simplifying the thresholded CT volume and standardizing the imaging data with denoising, we apply a series of morphological operations and median filtering. Specifically, we perform morphological closing on the input image using a disk with a fixed radius of $10$. The resulting image is then filtered using median filtering in a $3$x$3$ neighborhood. As the CT volume is affected by noise in the form of holes and intensity peaks, the pre-processing pipeline in our framework is meant to make the volume smooth for the next phase.

\subsection{Initial Boundary Estimation}
We identify the skin boundary of the subject by filling image regions and extract skin-boundary contour from the filled image. The largest set of contour points on the filled image corresponds to the skin boundary. We then generate a set of boundary points between the boundary contour and the centroid of the skin boundary serving as possible \textbf{hypotheses} for SAT-VAT boundary locations. Each hypothesis (candidate boundary location) is next verified for the possibility of being a boundary location by assessing gradient information. Let $H_{C}=\{h_1,h_2,h_3...h_n\}$ be the set of hypothesis points between the initial boundary contour $C$ and the center of the abdomen region $m$. Then, SAT-VAT boundary locations $B=\{b_1,b_2,b_3...b_n\}$ would satisfy the following condition:
\begin{equation}
		h_j \neq h_{j-1}       \quad \text{for } h_j \in B,  \text{ and }  b_i \in H_{C}, \forall i.
\end{equation}

\noindent For each boundary location in $C$, we obtain SAT-VAT separating boundary location $b_i$ from the transition of pixel intensities from 1 to 0 as on the thresholded image, where labels are from the initial segmentation of CT volume. These boundary points can be noisy and are often stuck inside the small cavities of the subcutaneous fat. Hence, we propose a two stage approach to smooth these noisy measurements.

\subsection{Outlier Removal Using Geometric MAD}
In the first stage of the outlier removal and smoothing, we apply median absolute deviation (MAD) on the distances between the abdominal (B) and the skin boundaries (P). The intuition behind this idea is that abdominal boundary contour would maintain a smoothly varying distance from the skin boundary. The outliers in subcutaneous and visceral cavities usually have abruptly changing distances from the skin boundary; hence, we compute the MAD \cite{leys2013detecting} for all boundary points to obtain a score for each point in the boundary being an outlier. The MAD coefficient $\Phi$ is given by:  
\begin{equation}
			\Phi = \frac{\left|(\left\|P-B\right\|_2)-med(\left\|P-B\right\|_2)\right|}{med\{\left|(\left\|P-B\right\|_{2})-med(\left\|P-B\right\|_{2})\right|\}},
		\label{eq:mad}
	\end{equation}
where \textit{med} is the median and the denominator of Eq. \ref{eq:mad} is median absolute deviation. Boundary locations having $\Phi$ greater than empirically selected threshold 2.5 are labeled as outliers and are removed from the boundary $B$. The MAD is found to be quite effective in outlier removal against noisy boundary measurements in our experiments. However, there may be still some boundary locations that could potentially lead to drifting of SAT-VAT separation, particularly in small cavities. To mitigate the influence of those boundary points on the final boundary $B$, we apply appearance constraints, explained in the next subsection.

\subsection{Outlier Removal Using Appearance Attributes}
\begin{figure}[t!]
		 \hspace*{-0.5cm}
	  \begin{subfigure}[b]{4.5 cm}
	   \includegraphics[height=4.0 cm, width=4.5cm]{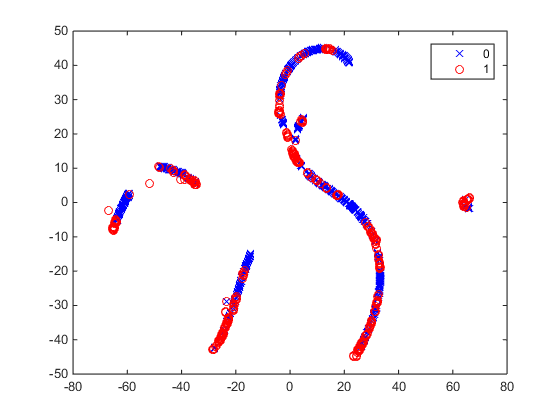}
	   \caption{}
	   \label{fig:eucdist}
	  \end{subfigure}%
		~
	  \begin{subfigure}[b]{4.5 cm}
	   \includegraphics[height=4.0 cm, width=4.5cm]{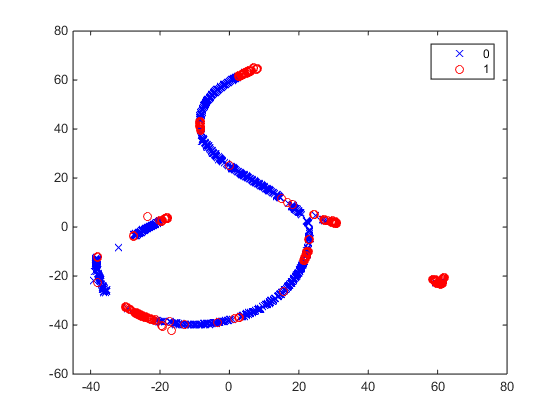}
	   \caption{}
	   \label{fig:corrdist}
	  \end{subfigure}%
	
	\caption{\textit{$t$-SNE visualizations using (a) Euclidean and (b) Normalized Correlation distance. Better separation between classes can be clearly seen in (b) where blue and red are two separate classes}}
		\label{fig:t-SNE}
		\vspace{-0.3cm}
	\end{figure}

For the second stage of the core methodology, we compute Histogram of Oriented Gradients (HOG) features as appearance attributes in a 14x14 cell with overlapping window of 5 pixels (i.e., resulting in a total of 279 dimensional feature vector). The goal is further identify the candidate boundary points between SAT and VAT where shape/geometry based attributes have limitations. In practice, as these boundary points lie on high dimensional manifold, we use \textbf{normalized correlation distance} instead of Euclidean distance, as also justified by computing the proximity $Q_{ij}$ between embedded boundary points $q_i$ and $q_j$ using $t$-distributed stochastic neighborhood embedding (t-SNE) \cite{t-SNE}:
\begin{equation}
Q_{ij} = \frac{1+(\left\Vert q_i-q_j\right\Vert_2)^{-1}}{\sum_{a \neq b}(1+(\left\Vert q_a-q_b\right\Vert_2)^{-1}}.
\end{equation}
Briefly, we use $t$-SNE to project the extracted HOG features on to a 2-dimensional space. Figure \ref{fig:t-SNE} depicts the feature embedding visualization using $t$-SNE, where better separation of features with normalized correlation distance can be readily observed. We then compute the local outlier scores $\Pi$, to get the confidence of each point being an outlier \cite{loop}. The intuition is to cluster points, that are mapped to denser regions in high dimensional feature space, together and to alienate the outliers which actually don't constitute the muscle boundary between SAT and VAT. The higher the $\Pi$, the higher the confidence of that boundary point being an outlier:
\begin{equation}
\Pi(x)=erf \left ( \frac{PLOF(x)}{\sqrt{2} . nPLOF} \right ),
\end{equation}
where $erf$ is the Gaussian Error Function and PLOF is the probabilistic local outlier factor based on the ratio of the density around point $x$ and the mean value of estimated densities around all remaining points. nPLOF is the $\lambda$ standard deviation of the PLOF, where $\lambda=3$ in our experiments. 

\subsection{Context Driven Label Fusion}
We employ Conditional Random Fields (CRF) \cite{Fulkerson2009} to fuse the labels of the boundary candidates after the second and third stages, i.e., to remove any inconsistencies in geometric and appearance labels using context information. In CRF formulation, the image is considered as a graph G=(V,E), where the nodes (V) consists of only the hypothesis boundary points and the edges (E) are neighboring points in high dimensional feature space. We seek to minimize the negative log of $Pr(k|G;w)$ with $k$ labels and weights $w$ as:
\begin{equation}
\hspace{-0.3cm}
-log(Pr(k|G;w))=\sum_{v_i\in V} \Theta (k_i|v_i) + w  \sum_{v_i,v_j\in E}\Psi (k_i,k_j|v_i,v_j).
\end{equation}

\noindent
\textit{Unary Potentials}\\
The unary potentials are defined to be probability outputs obtained after applying k-means clustering on the normalized scores of first and second stages:

\begin{table}[H]
\begin{center}
\vspace{0 cm}
\caption{\textit{
Segmentation results for SAT and VAT evaluated by Dice Similarity Coefficient (DSC)
}}\label{table:Results}\vspace{-0.03in}
\label{table:quanresults_dsc}
\footnotesize{
\begin{tabular}{l@{\hspace{0.25in}}c@{\hspace{0.25in}}c@{\hspace{0.1in}}c}
\toprule[1.5pt] \multirow{2}{*}{\textbf{Methods}}   & \multirow{2}{*}{\textbf{Subcutaneous DSC}}   &   \multirow{2}{*}{\textbf{Visceral DSC}} \\ 
\\
\cmidrule(r){1-4}
Zhao et al. \cite{zhao2006}    &      89.51\%      &   84.09\%   &        \\
RANSAC      &      91.14\%    &     85.90\% &     \\
Geometric MAD      &     89.61\%      &      87.58\%        \\
Appearance LoOS     &         92.45\%      &    88.48\%   &          \\
Context Driven Fusion      &     \textbf{94.04\%}     &   \textbf{91.57\%}   &         \\
\toprule[1.5pt]
\end{tabular}}
\end{center}\vspace{-0.2in}
\end{table}

\begin{equation}
\Theta (k_i|v_i)=-log(Pr(k_i|v_i))
\end{equation}

\noindent
\textit{Pairwise Potentials}\\
The pairwise potentials between the set of neighboring points $v_i$ and $v_j$ are defined as:
\begin{equation}
\Psi (k_i,k_j|v_i,v_j)=\left ( \frac{1}{1+\left | \phi _i-\phi _j \right |} \right )[k_i\neq k_j],
\end{equation}
where $\left | . \right |$ is the Manhattan or $L_1$ distance and  $\phi$ is the concatenation of appearance and geometric features.

Finally, we fit a convex-hull around the visceral boundaries obtained after the label fusion stage. The segment inside the convex-hull is masked as VAT and that outside of it is labeled as SAT.

\vspace{-0.4 cm}
\section{Results}
\vspace{-0.2 cm}
\textbf{Data:} With IRB approval, we retrospectively collected imaging data from \textbf{151} oncology patients who underwent PET/CT scanning in our institute from 2011 to 2013 (67 men, 84 female, mean age: 57.4). Since CT images are from PET/CT hybrid counterpart; they are in low resolution, and no contrast agent was used.  In-plane resolution ($xy$-plane) of CT image was recorded as 1.17 mm by 1.17 mm, and slice thickness was 5 mm. CT images were reconstructed on abdominal level before applying the proposed method. Patients were selected to have roughly equal distribution from varying BMI metrics (obese, overweight, normal, and underweight). Two expert interpreters manually labeled the whole data set to serve as ground truth. Above 99\% of agreement was found with no statistical difference between observers' evaluations (p >0.5).
\\
\begin{table}[H]
\centering
\caption{\textit{Mean absolute error (MAE) for SAT and VAT in ml.}}
\footnotesize{
\begin{tabular}{|l||l|l|ll}
\cline{1-3}
\textbf{Method} & \textbf{MAE for SAT (ml)} & \textbf{MAE for VAT (ml)} &  &  \\ \cline{1-3} 
RANSAC &   \multicolumn{1}{c|}{12.4995}    &    \multicolumn{1}{c|}{12.5009}   &  &  \\ \cline{1-3}
Zhao et al. \cite{zhao2006} &  \multicolumn{1}{c|}{11.1617}      &  \multicolumn{1}{c|}{11.1632}     &  &  \\ \cline{1-3}
Geometric MAD &       \multicolumn{1}{c|}{13.6875}      &    \multicolumn{1}{c|}{13.6890}      &  &  \\ \cline{1-3}
Appearance LoOS &      \multicolumn{1}{c|}{11.8463}      &    \multicolumn{1}{c|}{11.8473}   &  &  \\ \cline{1-3}
Context Driven Fusion & \multicolumn{1}{c|}{\textbf{7.1258}}     & \multicolumn{1}{c|}{\textbf{7.1281}}   &  &  \\ \cline{1-3}
\end{tabular}}
\label{table:quan_mae}
\end{table}

\noindent \textbf{Evaluations:} For segmentation evaluation, we used widely accepted dice similarity coefficient (DSC): $\frac{2\left | I_G \cap I_S \right |} { \left | I_G \right| + \left| I_S \right|}$, where $I_G$, $I_S$ were ground truth and automatically segmented fat regions, respectively. For quantification evaluation, we used Mean Absolute Error (MAE) in milliliters (ml) of adipose tissue.\\

\begin{figure}[t!]
	  \begin{subfigure}[b]{4.5 cm}
	   \includegraphics[height=3.5 cm, width=4.0cm]{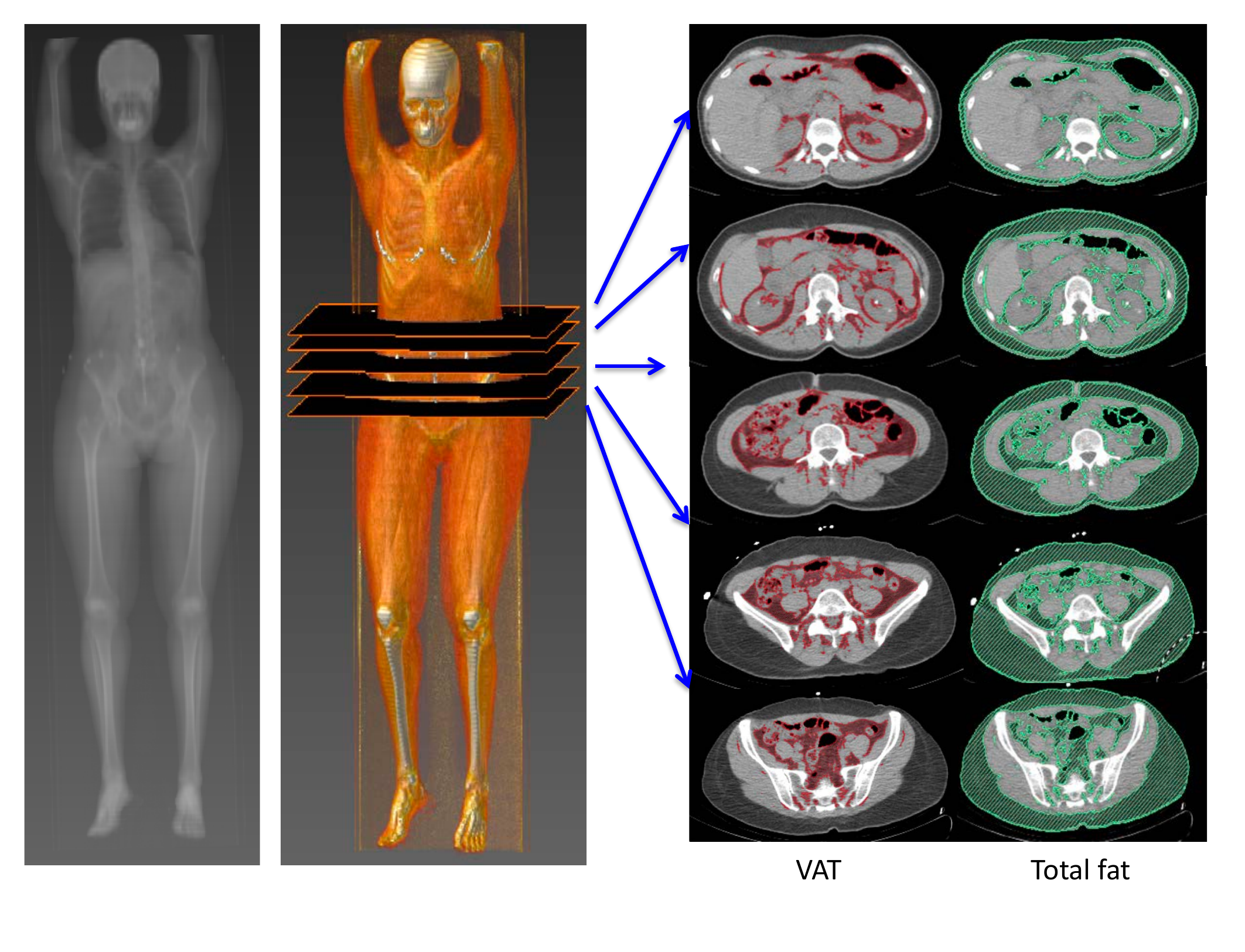}
	   \caption{}
	   \label{fig:DSC-SAT}
	  \end{subfigure}%
		~
	  \begin{subfigure}[b]{4.5 cm}
	   \includegraphics[height=3.5 cm, width=4.0cm]{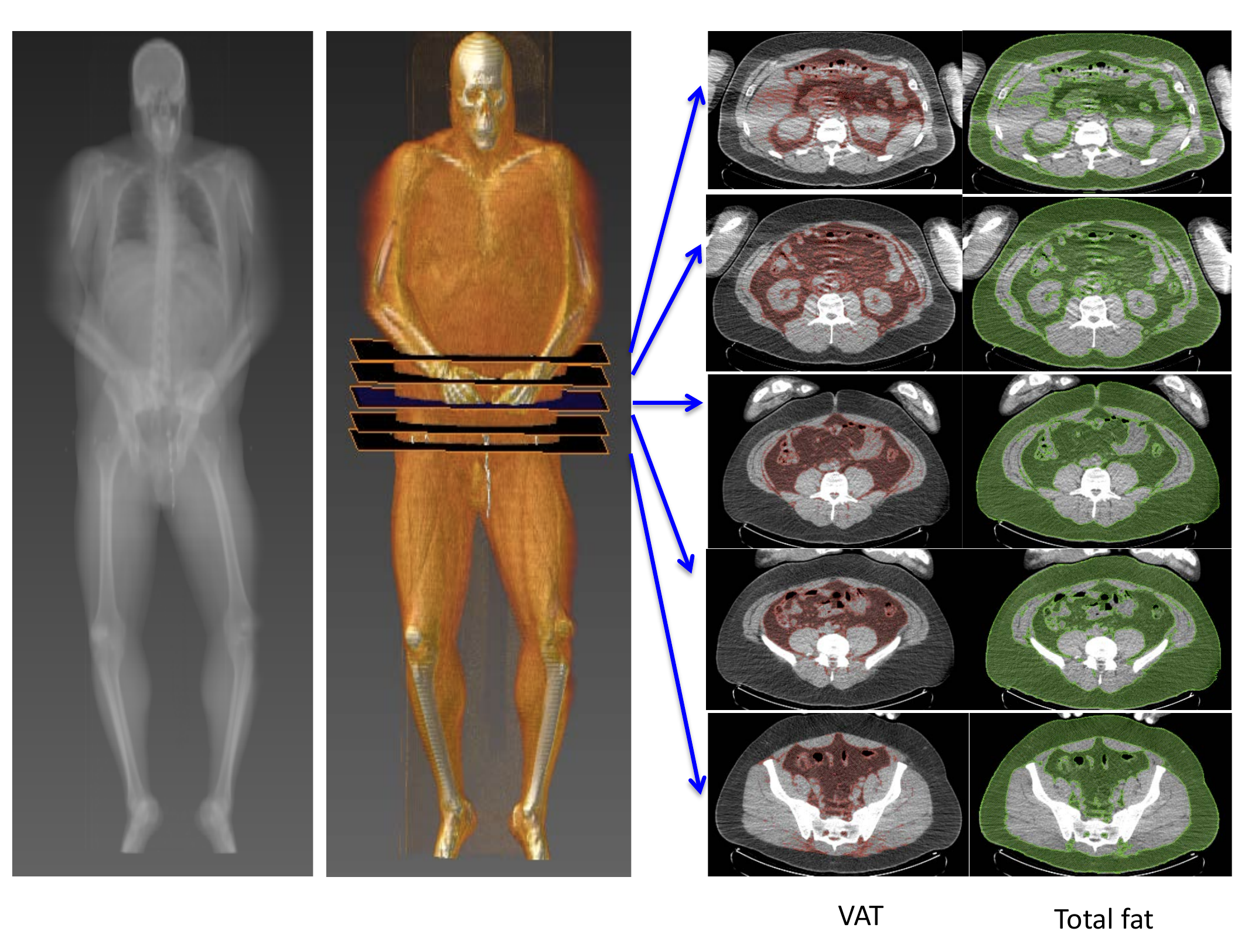}
	   \caption{}
	   \label{fig:DSC-VAT}
	  \end{subfigure}%
	  \vspace{-0.3cm}
		\caption{\textit{Volume rendering along with labeled VAT and total fat in abdominal region is seen. (a) Subject with BMI$\textless$25, (b) subject with BMI$\textgreater$30. VAT (red) is shown to be separated from total fat (green). Five slices at the umbilical level of the abdomen are shown for illustration.}}
		\label{fig:Volrend}
		\vspace{-0.4 cm}
	\end{figure}

\noindent \textbf{Comparisons:} We compare our proposed method with Zhao et al. \cite{zhao2006} and RANSAC. Also, we have progressively shown the results of our proposed framework's steps: Geometric MAD, Appearance LoOS and the final context driven fusion. DSC results for SAT and VAT using these five methods are shown in Table \ref{table:quanresults_dsc} where significant improvement can be seen using our proposed method. Moreover, Table \ref{table:quan_mae} shows a comparison using mean absolute error (MAE) in ml, where the proposed method records $>$36\% lesser MAE as compared to \cite{zhao2006} and other baselines. Figure \ref{fig:Volrend} shows the volume rendering of subjects with BMIs in normal and obese range along with visceral and total fat segmentations.\\

\noindent \textbf{Computation Time:} The unoptimized MATLAB implementation of the proposed method takes around 2.59 seconds per slice on Intel Xeon Quad Core CPU @ 2.80GHz and 24.0GB RAM. For faster performance the algorithm can run in parallel for multiple slices as there are no dependencies across slices.

	\vspace{-0.2 cm}
\section{Conclusion}
\vspace{-0.3 cm}
In this work, we present a novel approach for an automatic and unsupervised segmentation of SAT and VAT using global and local geometric and appearance attributes followed by context driven label fusion. Evaluations were performed on low resolution CT volumes from 151 subjects with varying BMI conditions to avoid bias in our analysis. The proposed method has shown superior performance as compared to other methods. As an extension of this study, we are currently developing an automated abdominal region detection algorithm to extract abdominal region from the whole body CT scans and combine it with the proposed framework for fat quantification to be used in routine clinics.

Our approach is designed as model free; training or patient specific parameter tuning is not necessary. Instead, image-driven features are adapted accordingly due to the robustness of our proposed method. This study is an important first step in generalizing fat quantification using low resolution CT images, which will be more likely in the near future due to an exponential increase in the quantity of publicly available CT data. 
\bibliographystyle{IEEEbib}
\bibliography{root}

\end{document}